\documentclass{article}
\usepackage{amsmath, amssymb, amsthm}
\usepackage[margin=1in]{geometry}
\usepackage{graphicx}
\usepackage{booktabs}
\usepackage{tikz-cd}

\theoremstyle{definition}
\newtheorem{definition}{Definition}
\theoremstyle{plain}
\newtheorem{theorem}{Theorem}
\newtheorem{lemma}{Lemma}
\newtheorem{proposition}{Proposition}

\theoremstyle{remark}
\newtheorem{remark}{Remark}

\DeclareMathOperator*{\argmin}{\arg\!\min}

\title{Learning From Simulators: A Theory of Simulation-Grounded Learning}
\author{Carson Dudley, Marisa Eisenberg}
\date{\today}

\begin{document}
\maketitle

\begin{abstract}

Simulation-Grounded Neural Networks (SGNNs) are predictive models trained entirely on synthetic data from mechanistic simulations. They have achieved state-of-the-art performance in domains where real-world labels are limited or unobserved, but lack a formal underpinning.

We place SGNNs in a unified statistical framework. Under standard loss functions, they can be interpreted as amortized Bayesian predictors trained under a simulator-induced prior. Empirical risk minimization then yields convergence to the Bayes-optimal predictor under the synthetic distribution. We employ classical results on distribution shift to characterize how performance degrades when the simulator diverges from reality. Beyond these consequences, we develop SGNN-specific results: (i) conditions under which unobserved scientific parameters are learnable via simulation, and (ii) a back-to-simulation attribution method that provides mechanistic explanations of predictions by linking them to the simulations the model deems similar, with guarantees of posterior consistency.

We provide numerical experiments to validate theoretical predictions. SGNNs recover latent parameters, remain robust under mismatch, and outperform classical tools: in a model selection task, SGNNs achieve half the error of AIC in distinguishing mechanistic dynamics. These results establish SGNNs as a principled and practical framework for scientific prediction in data-limited regimes.
\end{abstract}

\section{Introduction}

A central challenge in scientific machine learning is to construct predictive models in domains where real-world data are scarce, noisy, or unobservable. In public health, for example, accurate estimates of the number of cases in an infectious disease outbreak may not be available until late in the outbreak. In ecology, essential quantities like species population limits in an environment may not be directly observable from collected data. These limitations render standard supervised learning approaches ineffective: when data are limited, empirical learners fail to generalize; when targets are unobservable, learning is altogether impossible.

Simulation-Grounded Neural Networks (SGNNs) were recently developed by our group as a class of predictive models for scientific domains where real-world data are limited or unobservable \cite{sgnns}. Rather than training on empirical observations, SGNNs learn from synthetic datasets generated by mechanistic simulations diversified across parameters, model structures, stochastic processes, and observational artifacts. This paradigm has already yielded state-of-the-art performance across a range of prediction tasks. During early COVID-19, SGNNs nearly tripled forecasting skill relative to the CDC’s median model, despite never being trained on real-world data \cite{mantis}. In ecology and chemistry, they outperform domain-specific baselines, reduce variance in high-dimensional time-series forecasting, and improve chemical yield prediction with only minutes of pretraining \cite{sgnns}.

Despite these empirical successes, the theory of SGNNs remains underdeveloped. When and why do SGNNs work? What function classes become learnable under simulation that are unlearnable from empirical data? How does performance degrade when simulations diverge from reality? And can simulation-trained models provide principled, mechanistic interpretations of their predictions?

To address these questions, we draw on two complementary literatures: classical statistical learning theory and model identifiability theory. From the former, we adapt classic results on generalization and distribution shift to the SGNN setting, clarifying how performance on synthetic data translates to guarantees under simulator–reality mismatch \cite{mohri2018foundations}. From the latter, we formalize when unobserved parameters (e.g., the transmissibility of a disease) are estimable through simulation \cite{identifiability1, identifiability2, identifiability3}, introducing a class of simulation-learnable functions. This intersection highlights both the opportunities and the limits of simulation-grounded learning.

This paper develops a mathematical theory of simulation-grounded prediction. We introduce a formal framework in which SGNNs are cast as amortized Bayesian estimators under a simulator-induced prior. Our key results include:
\begin{itemize}
\item \textbf{Theory of SGNNs.} We define Simulation-Grounded Neural Networks (SGNNs) as amortized Bayesian predictors trained over a simulator-induced prior, and prove that they converge to the Bayes-optimal predictor under the synthetic data distribution.
\item \textbf{Generalization under misspecification.} We characterize how SGNNs perform when the simulator does not perfectly match reality, showing that test-time error separates into a learnable component and a penalty from simulator–reality mismatch. This framing offers practical guidance for evaluating models and improving simulator design.
\item \textbf{Learning unobservable quantities.} We prove that SGNNs can consistently learn scientific quantities that are unobserved in real-world data, such as the basic reproductive number of a disease, carrying capacity of a population, or source of information diffusion, under standard identifiability conditions.
\item \textbf{Mechanistic interpretability.} We formalize back-to-simulation attribution, a framework for explaining predictions in terms of the latent simulator parameters that generated them. We prove that attribution yields consistent estimates of the underlying generative mechanisms, and show that with a lightweight alignment objective, it recovers the full posterior distribution over simulator parameters, enabling simulation-grounded explanations and uncertainty quantification.
\item \textbf{Empirical validation.} We confirm theoretical predictions using controlled simulators, demonstrating convergence to Bayes predictors, graceful degradation under mismatch, and accurate recovery of latent parameters via attribution.
\end{itemize}

These results position SGNNs as a theoretically grounded framework for prediction in data-limited scientific settings. They identify a class of simulation-learnable functions, clarify how simulation-grounded learning generalizes under misspecification, and provide guarantees for both inference and mechanistic interpretability. \footnote{{All code for experiments in this paper is available at \texttt{https://github.com/carsondudley1/SGNNtheory}}.}

\subsection*{Related Prior Work}

In simulation-based inference (SBI) and likelihood-free inference (LFI) \cite{cranmer2020frontier}, neural networks are used to approximate statistical functions like posterior distributions, likelihoods, or likelihood ratios to enable parameter inference. SGNNs take a different approach: they learn task-specific predictors directly via supervised training on synthetic data, whether the task is parameter inference, future prediction, or classification.

SGNNs also relate to physics-informed neural networks (PINNs) \cite{raissi2019pinns}, neural operators \cite{neuraloperator}, and hybrid mechanistic–machine learning models \cite{uodes}, all of which integrate scientific knowledge into model architectures, optimization constraints, or inductive biases. These approaches typically assume access to labeled real-world data and use mechanistic priors to improve generalization. In contrast, SGNNs embed domain knowledge into the training data itself: they rely entirely on simulation and are designed to operate even when labeled observations are unavailable or unobservable.

The simulation-grounded learning paradigm has emerged across scientific domains without a unified theoretical framework. In epidemiology, for instance, early work like DEFSI trained neural networks on agent-based mechanistic simulations for influenza forecasting. \cite{defsi} A similar practice known as ``sim2real'' is common in robotics, where models are trained or evaluated in simulated environments and deployed on real systems \cite{sim2real}. The SGNN concept unifies these domain-specific applications under a single paradigm, and this paper provides a formal theoretical grounding.

The closest conceptual relative is the class of Prior-Data Fitted Networks (PFNs) \cite{pfn, tabpfn, nagler2023foundations}, which learn amortized Bayesian predictors via training on synthetic tasks sampled from a prior. SGNNs extend this idea to scientific domains, where the prior is defined implicitly by a mechanistic simulator. This induces structured, non-exchangeable data with latent parameters and domain-specific noise, and allows SGNNs to learn mappings that are grounded in physical or biological processes.

We quantify error under model misspecification and prove consistency of simulation-based attribution mechanisms. To do so, we draw from classical learning theory, particularly in the study of excess risk and generalization under distribution shift \cite{mohri2018foundations}.

\section{Definitions and Formal Setup}

We formalize simulation-grounded learning as a framework in which predictive models are trained exclusively on synthetic data generated from a structured scientific simulator. This section introduces the key mathematical objects, generative assumptions, and notation that underlie our theoretical results.

\subsection{Simulators as Generative Models}

Let $\Theta$ denote a space of latent scientific configurations, including both parameters (e.g., reaction constants in chemistry) and structural choices (e.g., model type, presence of mechanisms). Each $\theta \in \Theta \subseteq \mathbb{R}^d$ parameterizes a \emph{scientific simulator}, which we define as the composition of two components:

\begin{itemize}
  \item A \textbf{mechanistic model} $\mathcal{M}(\theta)$, which governs the latent system dynamics (e.g., transmission processes, ecological interactions, chemical reactions). This model typically consists of differential equations, discrete stochastic rules, or agent-based systems.
  \item An \textbf{observation model} $\mathcal{O}$, which maps latent system states to observed quantities, introducing realistic noise, reporting artifacts, or partial observation (e.g., delays, rounding, censoring).
\end{itemize}

\begin{definition}[Mechanistic Simulator]
Let $\theta \in \Theta \subseteq \mathbb{R}^d$ denote a vector of simulator parameters, which may include both continuous and discrete components. The mechanistic simulator $\mathcal{S}$ is defined as the composition:
\[
\mathcal{S} := \mathcal{O} \circ \mathcal{M},
\]
where $\mathcal{M}(\theta)$ is a mechanistic model (e.g., differential equations, stochastic process) and $\mathcal{O}$ is an observation model that introduces noise, partial observation, or reporting artifacts. For a given $\theta$, the simulator $\mathcal{S}(\theta)$ stochastically generates a sample $(x, y) \in \mathcal{X} \times \mathcal{Y}$, where $x$ is the observed input and $y$ is a task-specific target.
\end{definition}

We assume $\theta$ is drawn from a prior distribution $P(\theta)$ that encodes domain knowledge or structural uncertainty. The overall generative process is:
\[
\theta \sim P(\theta), \quad w \sim \mathcal{M}(\theta), \quad x = \mathcal{O}(w), \quad y = T(\theta),
\]
where $w$ denotes latent system trajectories and $T: \Theta \to \mathcal{Y}$ is a task-specific labeling function that defines the prediction target (e.g., future dynamics, class label, latent parameter). In some tasks, $T(\theta)$ may represent a deterministic label (e.g., a class label or model parameter), while in others (e.g., forecasting) it may be stochastic, depending on additional process noise shared with the input $x$. We allow for both cases, and in general treat $y$ as drawn from a conditional distribution $p(y \mid \theta)$.

\subsection{Synthetic Data and Simulation-Grounded Prediction}

Let $\mathcal{X}$ denote the space of observable inputs $x = \mathcal{S}(\theta)$, and let $\mathcal{Y}$ denote the space of targets $y = T(\theta)$. The simulator and labeling function together induce a joint distribution $D_{\text{syn}}$ over $\mathcal{X} \times \mathcal{Y}$. Then:
\[
x = \mathcal{S}(\theta), \quad y = T(\theta), \quad \text{so that } (x, y) \sim D_{\text{syn}}.
\]
Let $\mathcal{D}_{\text{syn}} = \{(x_i, y_i)\}_{i=1}^N$ be a dataset of $N$ independent samples from this distribution.

\begin{definition}[\textbf{Simulation-Grounded Neural Network (SGNN)}]
A Simulation-Grounded Neural Network is a predictor $f_\phi: \mathcal{X} \to \mathcal{Y}$ trained to minimize supervised loss on synthetic data:
\[
\min_{\phi} \; \mathbb{E}_{(x, y) \sim D_{\text{syn}}} \left[ \ell(f_\phi(x), y) \right],
\]
where $\ell: \mathcal{Y} \times \mathcal{Y} \to \mathbb{R}_{\ge 0}$ is a prescribed loss function (e.g., mean squared error, cross-entropy).
\end{definition}

In practice, we restrict attention to a model class $\mathcal{F}$ (e.g., neural networks of a given architecture), with SGNN predictors $f_\phi \in \mathcal{F}$.

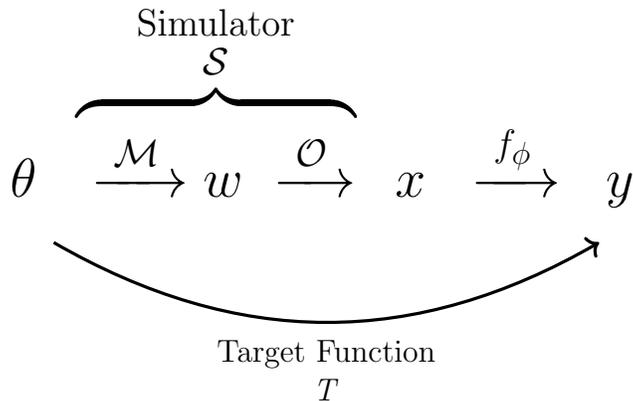
\begin{figure}
\begin{center}
{\huge
\[
\theta 
\;\;\overbrace{\rule{0pt}{3.5ex}\xrightarrow{\;\mathcal{M}\;} w \;\xrightarrow{\;\mathcal{O}\;}}^{\raisebox{0.5em}{$\substack{\text{Simulator} \\[3pt] \mathcal{S}}$}}\;\; 
x 
\;\;\xrightarrow{\;f_\phi\;}\;\; 
y
\]}
\vspace{-0.3em}
\begin{tikzpicture}
\draw[->, very thick] (-6.25,0) to[bend right=30] node[below=1mm] {$\substack{\text{\large Target Function} \\[3pt] \text{\large\textit{T}}}$} (1,0);
\end{tikzpicture}
\end{center}
\caption{\textbf{Simulation-grounded learning schematic.} Parameters $\theta$ sampled from a prior $P(\theta)$ are fed into a mechanistic model $\mathcal{M}$ (e.g., differential equations) to generate latent system dynamics $w$. An observation model $\mathcal{O}$ then transforms these into realistic observed data $x$ by adding noise, delays, and other artifacts. Together, $\mathcal{M}$ and $\mathcal{O}$ form the complete simulator $\mathcal{S} = \mathcal{O} \circ \mathcal{M}$. The SGNN $f_\phi$ learns to map from observations $x$ to target quantities $y = T(\theta)$, where $T$ can be simulator parameters, future trajectories, etc.}
\label{fig:schematic}
\end{figure}

Unlike empirical learners, which rely on real-world labeled data, SGNNs are trained entirely on simulated examples and are deployed directly on real data. Figure~\ref{fig:schematic} illustrates this core workflow: parameters $\theta$ sampled from a scientific prior drive mechanistic simulations to generate synthetic training pairs $(x,y)$, where $x$ represents realistic observations and $y = T(\theta)$ captures the target scientific quantity of interest.


\subsection{Model Misspecification and Real-World Generalization}

At test time, the SGNN is applied to inputs $x^*$ drawn from a true data-generating distribution $D_{\text{real}}$, which may differ from $D_{\text{syn}}$ due to modeling errors, structural misspecification, or prior mismatch.

\begin{definition}[Model Misspecification]
Model misspecification refers to the discrepancy between the synthetic distribution $D_{\text{syn}}$ and the real-world distribution $D_{\text{real}}$. This mismatch may arise from:
\begin{itemize}
  \item misspecified priors $P(\theta)$,
  \item missing mechanisms in the mechanistic model $\mathcal{M}$,
  \item or inaccuracies in the observation model $\mathcal{O}$.
\end{itemize}
\end{definition}

Quantifying the effect of model misspecification on SGNN performance is a primary goal of this paper.

\section{SGNNs as Amortized Bayesian Predictors}
\label{sec:sgns-bayes}

With the formal setup in place, we now recast simulation-grounded prediction as an instance of \emph{amortized Bayesian inference}. 

\paragraph{Amortized vs.\ standard inference.} 
In standard Bayesian inference, one computes a posterior distribution $p(\theta \mid x)$ separately for each new observation $x$, typically via sampling or optimization. This is accurate but computationally intensive: inference must be repeated from scratch for every input.  
Amortized inference, by contrast, learns a single global function $f_\phi(x)$ that maps any input $x$ directly to a posterior quantity of interest (e.g., an expectation or a parameter estimate). The cost of inference is ``amortized'' across many training examples: expensive computation is done once up front, after which predictions on new data are nearly instantaneous.

\paragraph{The Bayes-optimal predictor.}
Under the synthetic distribution $D_{\text{syn}}$, each training pair is generated by first drawing parameters $\theta \sim P(\theta)$, then producing input $x = \mathcal{S}(\theta)$ and target $y = T(\theta)$. The Bayes-optimal predictor is the function that, for each input $x$, outputs the expected target value under the posterior:
\[
f^{\star}(x) = \mathbb{E}_{\theta \sim P(\theta) \mid x}[T(\theta)] = \mathbb{E}[y \mid x].
\]
In words, $f^\star(x)$ is the predictor that minimizes expected loss: it matches the target $y$ \emph{in expectation}, given the information available in $x$.

\subsection{The Bayes-Optimal Predictor under the Simulation Distribution}

Intuitively, the Bayes-optimal predictor is the function that makes the fewest mistakes on average: for any input $x$, it outputs the conditional expectation of the target $y$ given $x$.  
Formally, let $\ell:\mathcal{Y}\times\mathcal{Y}\to\mathbb{R}_{\ge0}$ be a loss function, and define the population risk of a predictor $f$ under the synthetic distribution as
\[
R_{\text{syn}}(f) = \mathbb{E}_{(x,y)\sim D_{\text{syn}}}\bigl[\ell(f(x),y)\bigr].
\]
When the loss is squared error, $\ell(f(x),y) = \|f(x)-y\|^{2}$, the unique minimizer of this risk is
\[
f^{\star}(x) = \mathbb{E}[\,y \mid x\,].
\]
Our central question is whether an SGNN trained on samples from $D_{\text{syn}}$ can approximate this Bayes-optimal predictor, and under what conditions it converges to it.

\subsection{SGNNs as Approximators of the Bayes-Optimal Predictor}

An SGNN \( f_\phi \) is trained to minimize empirical risk over a synthetic dataset drawn from \( D_{\text{syn}} \). A central question is how the expected performance of this learned predictor, \( R_{\text{syn}}(f_{\phi_N}) \), compares to the Bayes-optimal risk \( R_{\text{syn}}(f^{\star}) \), where $N$ denotes the size of the training dataset.

\subsubsection{Finite-Sample Generalization Bound}

The \emph{excess risk} of the learned model,
\[
R_{\text{syn}}(f_{\phi_N}) - R_{\text{syn}}(f^{\star}),
\]
can be decomposed into two core components:
\begin{enumerate}
    \item \textbf{Approximation error (\( \mathcal{E}_{\text{approx}} \)):} 
    The gap between the Bayes-optimal predictor and the best function within the chosen model class \( \mathcal{F} \). This reflects the expressiveness of the architecture.
    \item \textbf{Estimation error (\( \mathcal{E}_{\text{est}} \)):} 
    The gap between the empirical risk minimizer \( f_{\phi_N} \) and the best-in-class function, arising from training on a finite sample of size \( N \).
\end{enumerate}

This decomposition can be written formally as:
\[
R_{\text{syn}}(f_{\phi_N}) - R_{\text{syn}}(f^{\star}) =
\underbrace{\left( \inf_{f \in \mathcal{F}} R_{\text{syn}}(f) - R_{\text{syn}}(f^{\star}) \right)}_{\text{approximation error}}
+
\underbrace{\left( R_{\text{syn}}(f_{\phi_N}) - \inf_{f \in \mathcal{F}} R_{\text{syn}}(f) \right)}_{\text{estimation error}}.
\]

The decomposition of excess risk into approximation and estimation error is classical in statistical learning theory \cite{mohri2018foundations, bartlett2002rademacher}. The approximation error can be made arbitrarily small with a sufficiently rich model class (e.g., deep neural networks). The estimation error admits a finite-sample bound via generalization tools such as Rademacher complexity \cite{bartlett2002rademacher}.

\paragraph{Empirical-risk minimizer.}
To analyze the behavior of SGNNs, we consider the standard empirical risk minimization (ERM) framework. Assume the learned predictor \( f_{\phi_N} \) is obtained by minimizing empirical loss over a synthetic dataset \( \mathcal{D}_{\text{syn}} = \{(x_i, y_i)\}_{i=1}^N \) drawn i.i.d.\ from \( D_{\text{syn}} \):
\[
    f_{\phi_N} \in \arg\min_{f \in \mathcal{F}} \frac{1}{N} \sum_{i=1}^N \ell(f(x_i), y_i),
\]
where \( \mathcal{F} \) is the function class (e.g., a neural network architecture) and \( \ell \) is the loss function.

This raises a natural question: \emph{How does the performance of the finite-sample predictor \( f_{\phi_N} \) compare to the ideal Bayes predictor \( f^\star \)?}

We can adapt classic results in statistical learning theory to answer this qeustion \cite{mohri2018foundations, bartlett2002rademacher}, showing that the total excess risk decomposes cleanly into an approximation term (dependent on how expressive the model class is) and an estimation term (dependent on the dataset size \( N \) and complexity of \( \mathcal{F} \)).

\begin{theorem}[Finite-sample excess-risk bound]
\label{thm:finite-sample}
Let the loss $\ell:\mathbb{R}\times\mathcal{Y}\to[0,B]$ be convex in its first argument (a condition satisfied by standard choices such as mean squared error, cross-entropy, and quantile loss commonly used in SGNN training),
$L$-Lipschitz in that argument, and bounded by $B>0$. With probability at least $1-\delta$ over the i.i.d.\ draw of the synthetic dataset $\mathcal{D}_{\text{syn}}^{(N)}$, the ERM predictor $f_{\phi_N}$ satisfies:
\[
\boxed{%
R_{\text{syn}}(f_{\phi_N}) - R_{\text{syn}}(f^{\star})
\le
\underbrace{\inf_{f \in \mathcal{F}} R_{\text{syn}}(f) - R_{\text{syn}}(f^{\star})}_{\text{approximation error}}
+
\underbrace{4L \widehat{\mathfrak{R}}_N(\mathcal{F}) + 6B \sqrt{\frac{\log(2/\delta)}{2N}}}_{\text{estimation error}}}
\]
where \( \widehat{\mathfrak{R}}_N(\mathcal{F}) \) is the empirical Rademacher complexity of the function class \( \mathcal{F} \).
\end{theorem}

Proof deferred to Appendix~\ref{app:proofs}.

\paragraph{Interpretation.}
This bound shows that SGNNs trained via ERM will approach Bayes-optimal performance provided two conditions hold:
\begin{itemize}
    \item \textbf{Expressivity.} The hypothesis class $\mathcal{F}$ must be rich enough to approximate $f^\star$ (driving the approximation error toward zero).
    \item \textbf{Sample size vs.\ complexity.} As the number of synthetic samples $N$ grows, the estimation error decreases at a rate governed by the complexity of $\mathcal{F}$, here captured by the empirical Rademacher complexity $\widehat{\mathfrak R}_N(\mathcal{F})$. Larger classes need more samples to achieve the same accuracy.
\end{itemize}
In short, SGNNs converge to the Bayes predictor when the architecture is sufficiently expressive and trained on enough synthetic data relative to its capacity.
Both conditions are practical to satisfy in simulation-grounded learning: (1) large neural networks are universal approximators, and (2) synthetic data can be generated in arbitrarily large quantities. Together, this establishes a path to consistency: SGNNs can match the performance of the Bayes-optimal predictor given sufficient model capacity and synthetic training data.

\subsubsection{Consistency and Practical Considerations}

The finite-sample bound above shows that SGNNs trained via empirical risk minimization approach the Bayes-optimal predictor as two conditions are met: (1) the model class is sufficiently expressive, and (2) enough synthetic data is used. The following proposition formalizes this intuition by establishing consistency in the limit.

\begin{proposition}[Consistency of the \textsc{SGNN} Estimator]\label{prop:consistency}
Let the SGNN function class $\mathcal{F}$ be a universal approximator for the Bayes-optimal predictor $f^{\star}$, so that the approximation error vanishes: $\inf_{f \in \mathcal{F}} R_{\text{syn}}(f) = R_{\text{syn}}(f^\star)$. Further assume that the complexity of $\mathcal{F}$ is controlled such that the empirical Rademacher complexity vanishes with sample size:
\[
\lim_{N \to \infty} \widehat{\mathfrak{R}}_N(\mathcal{F}) = 0.
\]
Then, as the synthetic dataset size \( N \to \infty \), the ERM predictor \( f_{\phi_N} \) is consistent:
\[
\lim_{N \to \infty} R_{\text{syn}}(f_{\phi_N}) = R_{\text{syn}}(f^{\star}).
\]

Moreover, for squared-error loss, this convergence in risk is equivalent to convergence in the $L^2(\mathcal{D}_x)$ norm:
\[
R_{\text{syn}}(f) - R_{\text{syn}}(f^{\star}) = \mathbb{E}_{x \sim \mathcal{D}_x} \left[ \left(f(x) - f^{\star}(x)\right)^2 \right],
\]
which implies
\[
\lim_{N \to \infty} \mathbb{E}_{x \sim \mathcal{D}_x} \left[ \left(f_{\phi_N}(x) - f^{\star}(x)\right)^2 \right] = 0.
\]
\end{proposition}

\paragraph{Practical Optimization.}
The analysis above assumes that $f_{\phi_N}$ is the exact empirical risk minimizer. In practice, stochastic optimization yields an approximate solution $\tilde{f}_{\phi_N}$, introducing an additional \emph{optimization error}, $\mathcal{E}_{\text{opt}} = R_{\text{syn}}(\tilde{f}_{\phi_N}) - R_{\text{syn}}(f_{\phi_N})$. While typically small in modern training regimes, this third source of error contributes to the total excess risk. Regularization techniques (e.g., dropout, weight decay) also play a dual role by reducing estimation error via complexity control.

\begin{remark}[How SGNNs Achieve Bayesian Optimality]
SGNNs do not compute the posterior \( p(\theta \mid x) \) explicitly, nor do they perform traditional Bayesian inference. Instead, they achieve Bayes-optimal prediction through standard supervised learning.

The key mechanism is classical: the minimizer of mean squared error over a distribution is the conditional expectation of the target given the input (see, e.g., \cite{mohri2018foundations}). For simulation-grounded data, this conditional expectation is
\[
f^{\star}(x) = \mathbb{E}[y \mid x] = \mathbb{E}_{\theta \sim p(\theta \mid x)}[\mu(x, \theta)],
\]
where \( \mu(x, \theta) = \mathbb{E}[y \mid x, \theta] \) is the simulator's response.
Here the likelihood and posterior appear implicitly: the simulator defines a likelihood $p(x,y \mid \theta)$ through its mechanistic and observation components, and when combined with the prior $P(\theta)$ this induces the posterior $p(\theta \mid x)$.
Minimizing MSE on synthetic samples therefore coincides with learning the posterior expectation of $y$ given $x$, i.e., the Bayes-optimal predictor.

In short, the SGNN performs implicit Bayesian inference simply by fitting the right data with the right loss.
\end{remark}

\subsection{Experiment: SGNNs Approximate the Bayes-Optimal Predictor}
\label{subsec:bayes_validation}

We empirically validate that SGNNs perform amortized inference toward the Bayes-optimal predictor $f^*(x) = \mathbb{E}[\theta \mid x]$ under a known generative model. Specifically, we test whether an SGNN can recover the latent simulator parameters $\theta = (\alpha, \beta)$ from observed trajectories $x$, and compare its performance to a kernel-based Monte Carlo approximation of $f^*$.

\paragraph{Simulator.} We use a diagonal linear dynamical system (LDS) with additive Gaussian noise:
\[
x_{t+1} = A(\theta) x_t + \epsilon_t, \quad \epsilon_t \sim \mathcal{N}(0, \sigma^2 I), \quad x_0 = [1, 1]^\top,
\]
where the latent parameters $\theta = (\alpha, \beta)$ define
\[
A(\theta) = \begin{bmatrix} \alpha & 0 \\ 0 & \beta \end{bmatrix}, \quad \theta \sim \mathcal{U}([0.5, 1.5]^2).
\]
Each simulation yields a trajectory of length $T = 10$, flattened into a vector $x \in \mathbb{R}^{20}$. The target is the latent mechanism: $y = T(\theta) = \theta$.

Our analysis assumes the conditional mean $\mathbb{E}[y \mid x]$ is well defined, as is the case under Gaussian noise and other finite-variance distributions. For extremely heavy-tailed noise (e.g., Cauchy), the mean may not exist. In this setting, we can instead train with a robust convex loss (e.g., $\ell_1$ or Huber), in which case the Bayes predictor becomes the corresponding posterior median or quantile.

\paragraph{Task and Baseline.} 
The prediction task is to learn $f(x) \approx \mathbb{E}[\theta \mid x]$. 
We compare the SGNN to a nonparametric Monte Carlo estimator that reweights 10,000 reference samples with kernel-smoothed weights:
\[
\hat{f}^*(x) = \sum_i w_i(x) \theta_i, \quad 
w_i(x) \propto \exp\!\left(-\tfrac{\|x - x_i\|^2}{2\sigma^2}\right),
\]
with $\sigma^2$ set to the median pairwise distance between reference samples. 
This estimator provides a direct, nonparametric approximation to the Bayes predictor $f^\star(x)$ using only the simulator and prior. 
Unlike MCMC-based inference, it requires no additional modeling choices or convergence diagnostics, but it is bandwidth-sensitive, suffers from finite-sample noise, and must recompute weights for each new input. 
These limitations underscore the advantage of SGNNs: once trained, they perform amortized inference in a single forward pass.

\paragraph{SGNN Model.} We train an SGNN on $10^7$ synthetic pairs $(x, \theta)$ using mean squared error loss. The model amortizes inference, learning a global map from trajectories to latent parameters.

\paragraph{Evaluation.} On a 50,000-sample test set, we measure:
\begin{itemize}
    \item \textbf{Bayes convergence:} MSE between SGNN predictions and $\hat{f}^*(x)$.
    \item \textbf{Parameter recovery:} MSE between SGNN predictions and true parameters $\theta$.
\end{itemize}

\begin{figure}[ht]
    \centering
    \includegraphics[width=0.9\textwidth]{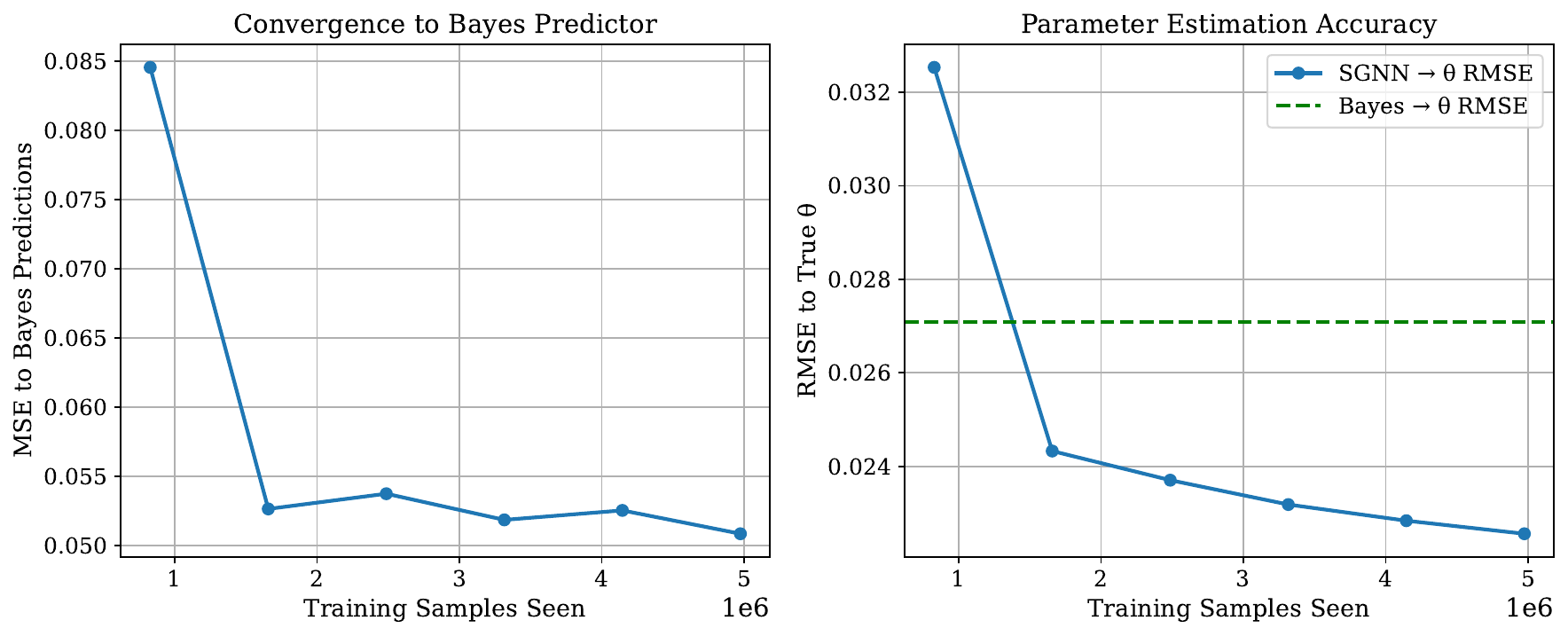}
    \caption{\textbf{SGNNs approximate the Bayes-optimal predictor.} \textbf{Left:} MSE between SGNN and Monte Carlo estimate of $f^*(x)$. \textbf{Right:} MSE between SGNN and ground-truth $\theta$. SGNNs outperform the baseline (dashed green line) due to amortized inference and smooth generalization.}
    \label{fig:sgnn_accuracy}
\end{figure}

\paragraph{Results.} Figure~\ref{fig:sgnn_accuracy} shows that SGNNs converge toward the Bayes-optimal predictor as training data increases (left). While the kernel estimator serves as our Monte Carlo approximation of $f^*$, it suffers from finite-sample variance and bias. The SGNN not only approaches this reference but achieves lower parameter estimation error than the kernel baseline (right), despite the latter's access to ground-truth simulator samples. This apparent outperformance arises because the SGNN amortizes inference across millions of samples, enabling smoother, more accurate generalization than the nonparametric estimator. These results demonstrate a practical advantage of SGNNs: in finite-sample settings, they can surpass standard approximators of the Bayes-optimal predictor by leveraging simulation-scale training and parametric inductive biases.

\section{Generalization Under Model Misspecification}
\label{sec:mismatch}

The preceding section established that, under the synthetic distribution $D_{syn}$, an SGNN converges to the Bayes-optimal predictor. We now quantify how its performance on a real-world test distribution, $D_{real}$, is affected by model misspecification, where $D_{real} \neq D_{syn}$. Our goal is to bound the excess risk of the SGNN on $D_{real}$ in terms of (i) its performance on the synthetic distribution and (ii) a formal divergence measuring the gap between the real and synthetic worlds.

\subsection{Formal Problem Setting}

\subsubsection{Distributions and Risks}
Let $D_{syn}$ be the synthetic distribution over pairs $(x,y)$ and let $D_{real}$ be the true data-generating distribution at test time. For a loss function $\ell:Y\times Y\to[0,L_{max}]$ bounded by $L_{max}$, we define the population risks as $R_{syn}(f) \triangleq \mathbb{E}_{(x,y)\sim D_{syn}}[\ell(f(x),y)]$ and $R_{real}(f) \triangleq \mathbb{E}_{(x,y)\sim D_{real}}[\ell(f(x),y)]$. The respective Bayes-optimal predictors are $f^{\star}_{syn} \triangleq \argmin_f R_{syn}(f)$ and $f^{\star}_{real} \triangleq \argmin_f R_{real}(f)$.

\subsubsection{SGNN Predictor}
The SGNN $f_{\phi_N}$ is the empirical risk minimizer over a synthetic training set $D_{syn}^{(N)}$ of size $N$:
\begin{align}
  \phi_N = \argmin_{\phi\in\Phi} \frac{1}{N} \sum_{(x_i,y_i)\in D_{syn}^{(N)}} \ell(f_\phi(x_i),y_i). \label{eq:erm-revised}
\end{align}
We assume the model family $\Phi$ has universal approximation capacity and that $f_{\phi_N}$ is a consistent estimator of $f^{\star}_{syn}$.

\subsection{Excess-Risk Bound under Mismatch}

Our goal is to control the real-world error in terms of the distance between $D_{syn}$ and $D_{real}$. To do so, we can apply classical distribution-shift inequalities to relate errors across $D_{syn}$ and $D_{real}$ \cite{mohri2018foundations}. We measure the discrepancy between distributions using the Total Variation (TV) distance, which is symmetric and well-suited for bounding differences in expectations.

\begin{definition}[Total Variation Mismatch]
The model misspecification is defined as the Total Variation distance between the real and synthetic joint distributions:
\[
  \Delta_{TV} \triangleq \sup_{A \subseteq \mathcal{X}\times\mathcal{Y}} |D_{real}(A) - D_{syn}(A)|.
\]
\end{definition}

\begin{theorem}[Generalization Bound for Model Misspecification]
\label{thm:mismatch}
The excess risk of the SGNN $f_{\phi_N}$ on the real distribution is bounded by the sum of its excess risk on the synthetic distribution and a penalty term proportional to the level of misspecification:
\[
  R_{real}(f_{\phi_N}) - R_{real}(f^{\star}_{real})
  \;\;\le\;\;
  \underbrace{
    R_{syn}(f_{\phi_N}) - R_{syn}(f^{\star}_{syn})
  }_{\text{Synthetic Excess Risk}}
  \;+\;
  \underbrace{
    2 L_{max} \Delta_{TV}
  }_{\text{Mismatch Penalty}}.
\]
\end{theorem}

Proof deferred to Appendix~\ref{app:proofs}.

\subsection{Interpretation and Practical Implications}

Theorem~\ref{thm:mismatch} provides a clear and powerful decomposition of the SGNN's test-time generalization error. The performance gap is attributable to two independent sources:

\begin{enumerate}
  \item \textbf{Synthetic Excess Risk:} This term, $R_{syn}(f_{\phi_N}) - R_{syn}(f^{\star}_{syn})$, measures how well the SGNN has learned to solve the task \emph{in the simulated world}. It is the sum of approximation error (is the model expressive enough?) and optimization error (did training find good parameters?). Under consistency assumptions, this term vanishes as the synthetic dataset size $N \to \infty$. It can be readily estimated using a held-out set of synthetic validation data.

  \item \textbf{Misspecification Penalty:} This term, $2 L_{max} \Delta_{TV}$, is a penalty determined solely by the intrinsic divergence between the simulator and reality. It is independent of the SGNN's architecture, training data volume, or optimization success. This bound highlights two key levers for improving real-world performance:
    \begin{itemize}
        \item \textbf{Improving the Simulator:} The primary way to reduce $\Delta_{TV}$ is by making the simulator more realistic. Increasing mechanistic diversity, adding observed real-world noise patterns, or refining the prior $P(\theta)$ all serve to close the gap between $D_{syn}$ and $D_{real}$, directly reducing this penalty.
        \item \textbf{Bounding Misspecification:} While $\Delta_{TV}$ is unknown, domain experts can often place a conservative upper bound on it by combining empirical checks (e.g., comparing simulated and real distributions of reporting delays, noise, or other observable artifacts) with scientific knowledge of plausible parameter ranges and mechanism validity. This provides a defendable estimate of the maximum simulator–reality gap. Practitioners can then calculate a safety margin for deployment by measuring the synthetic excess risk and adding this conservative misspecification penalty.
    \end{itemize}
\end{enumerate}

This framework clarifies that improving real-world performance requires both effective learning on the synthetic data (to reduce the first term) and high-fidelity simulation (to reduce the second).

\subsection{Estimating Misspecification in Practice}
\label{subsec:estimating_mismatch}

While the misspecification $\Delta_{TV}$ between the full joint distributions cannot be calculated directly (as $D_{real}$ is unknown), its value can be reasoned about by decomposing the sources of misspecification. This makes $\Delta_{TV}$ a powerful conceptual tool for model design and evaluation. Misspecification arises from three areas: the prior $P(\theta)$, the mechanistic model $\mathcal{M}$, and the observation model $\mathcal{O}$.

\begin{itemize}
    \item \textbf{Observation Misspecification ($\mathcal{O}$):} This is the most tractable component to estimate. We can often directly measure real-world observational artifacts. For instance, we can analyze a sample of real data to estimate the empirical distribution of reporting delays, the prevalence of weekend reporting dips, or the signal-to-noise ratio of a sensor. We can then quantify the divergence between these empirical distributions and their simulated counterparts, providing a lower bound on $\Delta_{TV}$.

    \item \textbf{Prior Misspecification ($P(\theta)$):} This reflects a mismatch in the assumed distribution of scientific parameters. While the true distribution is unknown, we can assess robustness through \emph{sensitivity analysis}. By systematically varying the prior $P(\theta)$ and observing the resulting change in $D_{syn}$, we can understand how sensitive our system is to prior assumptions. If $D_{syn}$ is stable under different plausible priors, the mismatch penalty is likely to be less severe. This can also be informed by consulting domain experts to ensure the support of $P(\theta)$ covers all scientifically plausible parameter regimes.

    \item \textbf{Mechanistic Misspecification ($\mathcal{M}$):} This is the most challenging source, since it includes both (i) uncertainty about which mechanisms already present in the simulator are essential, and (ii) the possibility that key mechanisms are entirely absent. For (i), ablation studies are informative: training models on simulators with and without particular mechanisms reveals which components are critical: if removing one causes a large drop in real-data performance, it must be retained to keep $\Delta_{TV}$ small. For (ii), missing mechanisms cannot be diagnosed by ablation alone. Instead, they may be detected through systematic residuals (where SGNN predictions consistently deviate from real data across simulations), domain expertise (experts know a process should exist, such as asymptomatic transmission), or failure modes (SGNN performs well in most regimes but breaks down in specific conditions). These signals indicate that the simulator family is insufficient and that $\Delta_{TV}$ is fundamentally bounded away from zero unless the missing mechanism is added.

\end{itemize}

\paragraph{Comparison with Traditional Mechanistic Modeling.}
A key difference from classical mechanistic modeling is how additional mechanisms are handled. In traditional modeling, lack of parsimony can cause model fitting to break down: over-specified models lead to unidentifiability, unstable inference, and inflated uncertainty, making it necessary to apply model selection criteria (e.g., AIC) or regularization methods (e.g., LASSO) to prune mechanisms \cite{aic, sindy, lassoode}. By contrast, SGNNs can be trained on a broad family of mechanisms simultaneously. As long as training covers the true mechanisms and sufficient synthetic data is generated, the network learns to prioritize the dynamics that matter for each individual case. This reduces the need for aggressive parsimony, shifting the focus from hand-selecting a minimal model to ensuring adequate mechanistic coverage in the simulator prior.

\subsection{Illustrative Cases of Misspecification}
\label{subsec:misspec_examples}

We now examine how the generalization bound in Theorem~\ref{thm:mismatch} manifests in practical settings. In particular, we decompose the total excess risk into the contributions of synthetic learning error and simulator–reality mismatch, and demonstrate how this decomposition provides theoretical insight into empirical failure modes.

\paragraph{Case 1: Robustness Under Slight Prior Misspecification.}
Assume the simulator encodes the correct generative mechanisms and noise profiles but assigns prior mass $P(\theta)$ that differs modestly from the true generative frequencies.

\begin{itemize}
\item \textbf{Mismatch Penalty:} Moderate but controlled. The support of $D_{\text{syn}}$ still substantially overlaps with that of $D_{\text{real}}$, though the mass is unevenly distributed.
\item \textbf{Synthetic Excess Risk:} Low. The SGNN successfully learns the Bayes-optimal predictor under $D_{\text{syn}}$.
\item \textbf{Implication:} The SGNN exhibits robust generalization, consistent with the bound in Theorem~\ref{thm:mismatch}. Because $f^*(x)$ integrates over the posterior $p(\theta \mid x)$, the learned function can extrapolate beyond high-density regions of the simulator prior, provided that those regions are not entirely excluded. This case illustrates robust generalization under mild mismatch, which we empirically validate in Section~\ref{subsec:empirical_mismatch} and which has been observed in practice \cite{mantis}.
\end{itemize}

\paragraph{Case 2: Overparameterized Simulator with Irrelevant Mechanisms.}
In an attempt to increase realism, the simulator includes a broad set of mechanisms, many of which do not affect the target task in practice.

\begin{itemize}
\item \textbf{Mismatch Penalty:} Low. Since the true data-generating process lies within the support of the synthetic distribution, the total variation distance $\Delta_{\text{TV}}$ remains small.
\item \textbf{Synthetic Excess Risk:} High. The task is more complex: identifying task-relevant structure requires disentangling signal from a large and potentially confounding mechanism space. For finite model capacity and training budget $N$, convergence to $f^*_{\text{syn}}$ may not be achieved.
\item \textbf{Implication:} The generalization bound is dominated by the excess synthetic risk. Even a ``correct'' simulator in the support sense may yield poor real-world performance if the induced prediction task is intractable. This highlights the dual requirement of representational sufficiency and statistical efficiency.
\end{itemize}

\paragraph{Case 3: Misspecification via Underparameterized Simulator.}
Consider a simulator that omits key stochastic or mechanistic components (e.g., generates noise-free data or ignores time-varying effects), or the simulator priors are too restrictive on the parameters, while the real world exhibits significant variability.

\begin{itemize}
\item \textbf{Synthetic Excess Risk:} Low. The SGNN achieves near-optimal performance on the simplified synthetic distribution; i.e., $f_{\phi_N} \approx f^*{\text{syn}}$.
\item \textbf{Mismatch Penalty:} High. The total variation distance $\Delta_{\text{TV}}$ between the real distribution $D_{\text{real}}$ and the synthetic distribution $D_{\text{syn}}$ is large, as critical real-world stochasticity is absent from the simulator.
\item \textbf{Implication:} The total generalization bound is dominated by the mismatch term. Although the SGNN performs well on synthetic data, it generalizes poorly due to a failure to account for real-world noise processes.
\end{itemize}

\paragraph{Case 4: Fundamental Misspecification.}
Here the simulator is not merely too simple, it is fundamentally incorrect. An essential mechanism is absent or the parameter bounds are signficantly incorrect, so the real data are generated by a process outside the simulator’s model class. Parameter diversity or prior tuning cannot fix a class error: the true distribution lies outside the support of $D_{\text{syn}}$, $\Delta_{TV}$ stays strictly positive, and the mismatch penalty persists. Remedy requires adding missing mechanisms or increasing parameter ranges (expanding the model class), not more data or different priors.
\begin{itemize}
\item \textbf{Synthetic Excess Risk:} Low, since the SGNN can still solve the task within the synthetic world.
\item \textbf{Mismatch Penalty:} Fundamentally high, since there is a systematic deviation between $D_{\text{syn}}$ and $D_{\text{real}}$.
\item \textbf{Implication:} Real-world performance will remain poor unless the simulator is corrected. Simulation-grounded learning cannot compensate for a fundamentally misspecified scientific model.
\end{itemize}

\paragraph{Summary.}
These examples clarify the distinct roles played by simulator coverage and induced task complexity. Theorem~\ref{thm:mismatch} not only offers a formal upper bound on real-world excess risk, but also provides actionable guidance for simulator design: models must balance fidelity to the true data-generating process with learnability of the induced prediction task.

\subsection{Empirical Validation of the Generalization Bound}
\label{subsec:empirical_mismatch}

We now empirically validate the generalization bound in Theorem~\ref{thm:mismatch} by constructing a controlled setting in which the degree of mismatch between the synthetic and real distributions can be precisely modulated. This allows us to examine how prediction error scales with distributional shift, and whether the observed degradation aligns with the theoretical guarantees.

\paragraph{Experimental Design.}
We use a linear dynamical system (LDS) as a generative model. Each trajectory is governed by the stochastic recurrence:
\[
    x_{t+1} = A x_t + \epsilon_t, \qquad \epsilon_t \sim \mathcal{N}(0, \sigma^2 I),
\]
where \( A \in \mathbb{R}^{d \times d} \) is a system matrix, and \( x_t \in \mathbb{R}^d \). We define a synthetic training distribution \( D_{\text{syn}} \) by fixing a reference matrix \( A_0 \), and define a perturbed test distribution \( D_{\text{real}} \) by introducing misspecification:
\[
    A^* = A_0 + \delta U,
\]
where \( U \) is a random matrix with unit Frobenius norm, and \( \delta \in \mathbb{R}_{\geq 0} \) controls the magnitude of deviation. This setup induces a continuum of test-time shifts, indexed by \( \delta \), while preserving the mechanistic structure of the simulator.

\paragraph{Learning Task.}
We train a neural predictor \( f_{\phi} \) on data drawn from \( D_{\text{syn}} \), where the task is to predict the next state \( x_{t+1} \) from the current state \( x_t \) under squared loss:
\[
    \ell(f_\phi(x_t), x_{t+1}) = \|f_\phi(x_t) - x_{t+1}\|^2.
\]
We then evaluate the trained predictor on data drawn from \( D_{\text{real}} \), for varying values of the mismatch parameter \( \delta \).

\paragraph{Theoretical Comparison.}
We evaluate whether the observed degradation in test performance is consistent with the two upper bounds implied by the generalization theory:
\begin{itemize}
    \item \textbf{Worst-case bound:} Based on total variation distance between the input-output distributions, approximated as \( \Delta_{\text{TV}} \le \delta \cdot \mathbb{E}[\|x_t\|] / \sigma \). For numerical stability, we upper bound \( \|x_t\| \le \sqrt{d} \).
    \item \textbf{Empirical data-dependent bound:} Computed as
    \[
    \Delta_{\text{TV}}^{\text{emp}} = \frac{1}{2\sigma} \cdot \mathbb{E}_{x_t \sim D_{\text{syn}}} \left[ \| (A^* - A_0)x_t \| \right],
    \]
    which tightens the worst-case bound using the actual observed displacement in predictions under mismatch.
\end{itemize}

\begin{figure}[ht]
    \centering
    \includegraphics[width=0.9\textwidth]{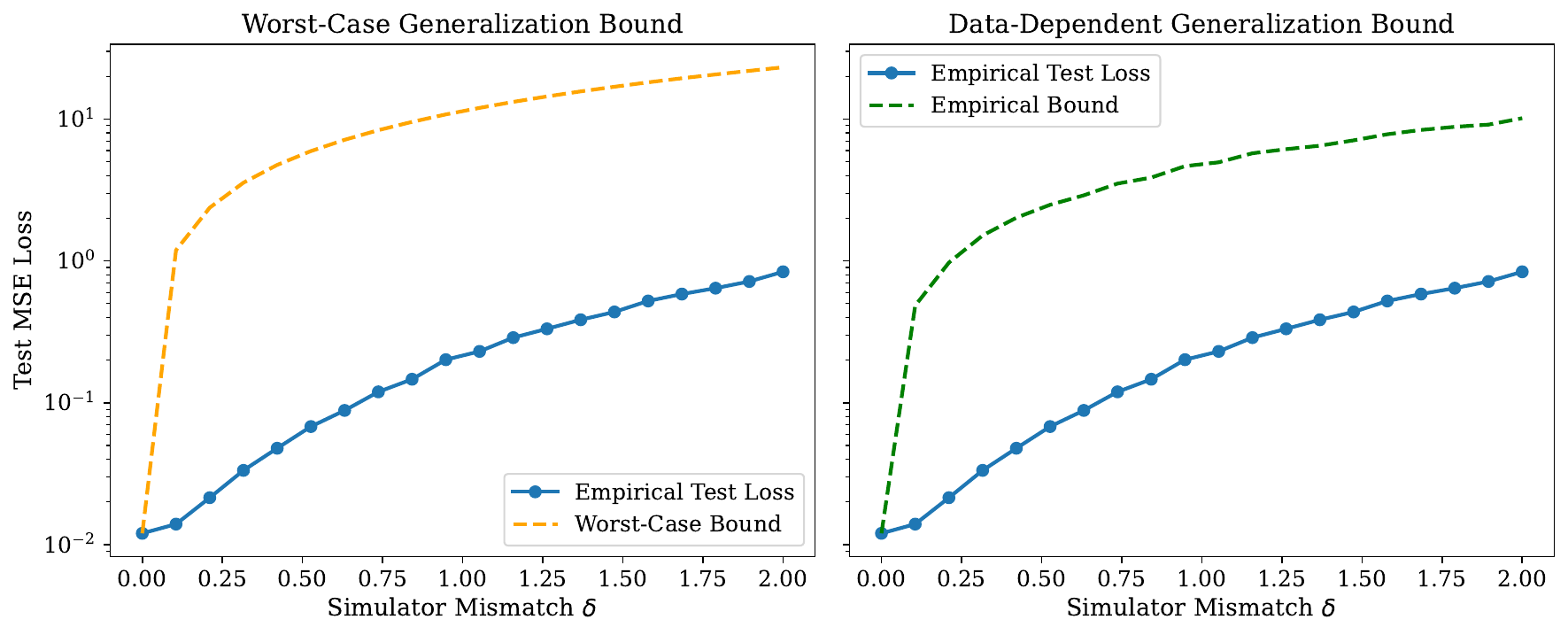}
    \caption{\textbf{Empirical validation of the SGNN generalization bound.} We plot the test loss of a predictor trained on synthetic data ($A_0$) and evaluated under increasing misspecification ($ A^* = A_0 + \delta U $). \textbf{Left:} Worst-case theoretical bound derived from total variation distance. \textbf{Right:} Empirical, data-dependent bound. In both cases, the actual test loss remains below the predicted generalization error, validating the theoretical guarantee in Theorem~\ref{thm:mismatch}.}
    \label{fig:generalization_bounds}
\end{figure}

\paragraph{Results.}
Figure~\ref{fig:generalization_bounds} plots the test loss of the SGNN predictor on \( D_{\text{real}} \) as a function of the mismatch level \( \delta \), alongside both the worst-case and empirical bounds. We observe:
\begin{itemize}
    \item The worst-case bound remains valid but conservative, as expected from its generality.
    \item The empirical bound tracks the test error more closely.
    \item The increase in test loss is smooth and monotonic with respect to $\delta$, consistent with the additive mismatch penalty predicted by Theorem~\ref{thm:mismatch}.
\end{itemize}

\section{Mechanistic Interpretability via Back-to-Simulation Attribution}
\label{sec:b2s}

Simulation-Grounded Neural Networks (SGNNs) enable not only accurate prediction, but also mechanistic interpretability by identifying which simulated mechanisms are most responsible for a given prediction. We formalize this process as back-to-simulation attribution and prove that, when trained with an appropriate alignment objective, it yields a consistent estimator of the posterior distribution over latent mechanisms.

\subsection{The Back-to-Simulation Attribution Framework}

Let $\theta \in \Theta \subset \mathbb{R}^d$ denote the latent simulator parameters, and let $x \sim p(x \mid \theta)$ denote simulated observations. Recall that the Bayes-optimal predictor of a target $T(\theta)$ is given by the posterior expectation:
\[
f^*(x) = \mathbb{E}[T(\theta) \mid x] = \int T(\theta) \, p(\theta \mid x) \, d\theta
\]

\textbf{The attribution question:} Given a trained SGNN's prediction $f_\phi(x)$ for some input $x$, which underlying mechanisms $\theta$ are most responsible for this prediction?

\textbf{Our approach:} We maintain a reference library $\mathcal{L} = \{(\theta_i, x_i)\}_{i=1}^M$ of simulator runs and their corresponding parameters. For any query input $x$, we:

\begin{enumerate}
\item \textbf{Encode} the query and library examples: $\phi(x), \phi(x_1), \ldots, \phi(x_M)$
\item \textbf{Compute similarity weights} based on embedding distance:
\[
w_i(x) = \frac{\kappa(\phi(x), \phi(x_i))}{\sum_{j=1}^M \kappa(\phi(x), \phi(x_j))}
\]
\item \textbf{Return the weighted distribution} over mechanisms: $\hat{p}_M(\theta \mid x) = \sum_{i=1}^M w_i(x) \delta_{\theta_i}$
\end{enumerate}

where $\kappa(\cdot, \cdot)$ is a similarity kernel (e.g., RBF kernel) and $\delta_{\theta_i}$ is a point mass at $\theta_i$.

\textbf{Intuition:} We attribute the prediction to mechanisms $\theta_i$ whose simulated outputs $x_i$ are similar (in embedding space) to our query $x$.

\subsection{Attribution Consistency Under Alignment Training}

The key insight is that attribution quality depends fundamentally on whether the learned embedding $\phi$ preserves the information needed to recover the posterior $p(\theta \mid x)$. Rather than hoping this happens by accident, we can explicitly train for it.

\begin{theorem}[Back-to-Simulation Attribution Consistency]
\label{thm:bsa_consistency}
Consider an SGNN trained with prediction loss $\mathcal{L}_{\text{pred}}$ and KL alignment loss:
\[
\mathcal{L}_{\text{KL}} = \mathbb{E}_{x \sim p(x)}\left[ \text{KL}\left(p(\theta \mid x) \,\|\, \sum_{i=1}^M w_i(x) \delta_{\theta_i} \right) \right]
\]

Assume: (1) Training converges: $f_\phi \to f^*$ and $\mathcal{L}_{\text{KL}} \to 0$, (2) Library samples $\{\theta_i\}$ become dense in $\Theta$, (3) $T(\theta)$ is continuous and bounded.

Then for any positive integer $k$:
\[
\sum_{i=1}^M w_i(x) T(\theta_i)^k \to \mathbb{E}[T(\theta)^k \mid x]
\]
In particular, the attribution distribution converges in all moments to the true posterior.
\end{theorem}

Proof deferred to Appendix~\ref{app:proofs}.

\textbf{Interpretation:} This shows that the attribution distribution $\sum_i w_i(x) \delta_{\theta_i}$ captures not only the mean of the true posterior (when $k=1$), but also its variance (when $k=2$), skewness, and all higher-order moments. This provides complete uncertainty quantification: the attribution weights correctly represent both the central tendency and the spread of plausible mechanisms.

\subsection{Comparison with Existing Interpretability Paradigms}
\label{subsec:interpretability_comparison}

Back-to-simulation attribution offers a fundamentally different perspective on interpretability compared to traditional methods. Rather than assigning importance scores to input features or referencing training examples, it provides mechanistic explanations in terms of the latent generative parameters that govern the underlying simulator.

\paragraph{Feature Attribution Methods.}
Standard approaches such as SHAP and LIME \cite{shap, lime} explain predictions by identifying the contribution of input features (e.g., pixels or covariates). These methods answer the question: “Which features of this specific input influenced the model’s prediction?”

By contrast, back-to-simulation attribution returns a distribution over latent parameters \( \theta \in \Theta \) such that the observed input \( x \) is likely under the simulation \( \mathcal{S}(\theta) \). This addresses a different and arguably more fundamental question: “What generative mechanism explains this observed behavior?”

\paragraph{Example-Based Methods.}
Influence functions and related techniques \cite{koh2020understanding} identify which training points most affected a given prediction. However, they remain tied to the training set and reflect empirical correlation rather than mechanistic causality.

In contrast, back-to-simulation attribution uses a reference library of simulated examples $\{(\theta_i, x_i)\}_{i=1}^M$, and retrieves examples not based on superficial similarity but shared generative origin. These attributions recover abstract mechanisms—even when corresponding trajectories differ substantially in their surface features.

\begin{center}
\begin{tabular}{ll}
\toprule
\textbf{Method} & \textbf{Explanatory Focus} \\
\midrule
SHAP, LIME & Input feature importance \\
Influence functions & Individual training point relevance \\
Back-to-simulation attribution & Latent scientific mechanism (\( \theta \)) \\
\bottomrule
\end{tabular}
\end{center}

\paragraph{Scientific Utility.}
Because SGNNs are trained on synthetic data generated from known mechanisms, back-to-simulation attribution yields explanations grounded in domain-relevant quantities. This makes it particularly well-suited for scientific machine learning, where understanding system behavior is often more important than maximizing predictive accuracy alone.

\subsection{Empirical Validation of Attribution Consistency}
\label{subsec:attribution_experiment}

We empirically validate Theorem~\ref{thm:bsa_consistency}, which establishes that back-to-simulation attribution with KL alignment training provides consistent estimators of all moments of the posterior distribution $p(\theta \mid x)$. Our experiments demonstrate that SGNNs trained with explicit attribution alignment can accurately recover both the mean and higher-order statistics of the generative mechanisms responsible for observed trajectories.

\subsubsection{Experimental Setup}

We simulate trajectories from a classical SIR (Susceptible-Infected-Recovered) model governed by two latent parameters: transmission rate $\beta$ and recovery rate $\gamma$. Parameters are drawn from independent uniform priors:
\[
\beta \sim \mathcal{U}(0.1, 0.5), \qquad \gamma \sim \mathcal{U}(0.05, 0.2)
\]

Each trajectory is a 50-step time series, from which the first 40 steps are used as input $x_{1:40}$, and the final 10 steps form the prediction target $x_{41:50}$. 

\textbf{Training with KL alignment:} Following Theorem~\ref{thm:bsa_consistency}, we train the SGNN with two objectives:
\begin{align}
\mathcal{L}_{\text{pred}} &= \mathbb{E}_{\theta \sim p(\theta)}\left[ \|f_\phi(x_{1:40}) - x_{41:50}\|^2 \right] \\
\mathcal{L}_{\text{KL}} &= \mathbb{E}_{x \sim p(x)}\left[ \text{KL}\left(p(\theta \mid x) \,\|\, \sum_{i=1}^M w_i(x) \delta_{\theta_i} \right) \right]
\end{align}

The total loss is $\mathcal{L}_{\text{total}} = \mathcal{L}_{\text{pred}} + \lambda \mathcal{L}_{\text{KL}}$ with $\lambda = 0.1$.

\subsubsection{Attribution Procedure}

We construct a reference library $\mathcal{L} = \{(\theta_i, x_i)\}_{i=1}^M$ of $M = 2000$ synthetic trajectories. For any test input $x^{\text{test}}$, attribution proceeds by:

\begin{enumerate}
\item \textbf{Encode} the query and library examples using the learned encoder $\phi$
\item \textbf{Compute similarity weights:}
\[
w_i(x^{\text{test}}) = \frac{\exp\left(-\frac{\|\phi(x^{\text{test}}) - \phi(x_i)\|^2}{h^2}\right)}{\sum_{j=1}^M \exp\left(-\frac{\|\phi(x^{\text{test}}) - \phi(x_j)\|^2}{h^2}\right)}
\]
\item \textbf{Return attribution distribution:}
\[
\hat{p}_M(\theta \mid x^{\text{test}}) = \sum_{i=1}^M w_i(x^{\text{test}}) \delta_{\theta_i}
\]
\end{enumerate}

This attribution distribution can then be used to compute any desired statistics, such as the posterior mean $\sum_i w_i(x) \theta_i$, posterior variance, or to assess uncertainty over the generative mechanisms.

\subsubsection{Results}

\begin{figure}[ht]
    \centering
    \includegraphics[width=0.6\textwidth]{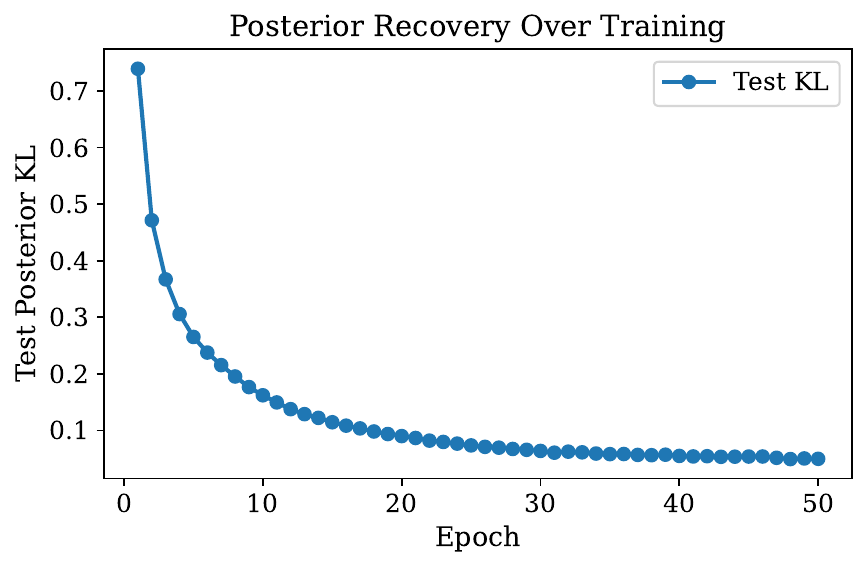}
    \caption{\textbf{Attribution consistency under KL alignment training.} KL divergence between SGNN attribution distribution and true posterior decreases over training epochs.}
    \label{fig:posterior_kl}
\end{figure}

\textbf{KL Alignment Convergence:} Figure~\ref{fig:posterior_kl} shows that the KL divergence between the SGNN attribution distribution and the true posterior decreases significantly during training, reaching values as low as 0.001--0.01. This confirms that the KL alignment loss successfully drives the attribution weights to approximate the true posterior $p(\theta \mid x)$.

\section{Learning Unobservable Scientific Quantities}
\label{sec:unobservable}

A core limitation of traditional supervised learning is its dependence on ground-truth labels.  
Many quantities scientists care about are fundamentally unobservable: e.g., a disease’s basic reproduction number ($R_0$), the mean number of people one infected individual will infect, or the mechanistic form governing an outbreak (e.g., whether there is asymptomatic transmission of the disease).  
SGNNs overcome this barrier by generating synthetic data in which such latent objects are known by construction, thereby turning an otherwise unlabelled problem into a supervised one.

\subsection{Formal Setting}

Let  
\[
\mathcal{M}\in\{\mathcal{M}_1,\dots,\mathcal{M}_K\},\qquad
\theta\in\Theta,\qquad
x=(\mathcal{O}\circ\mathcal{M})(\theta)
\]
denote, respectively, a mechanistic model, its latent parameters, and the resulting observable.  
We distinguish two target types:
\begin{enumerate}
    \item \textbf{Parametric target:} $y_{\text{param}} = T(\theta)$, where $T:\Theta\!\to\!\mathcal{Y}$ maps parameters to a scientific quantity (e.g.\ $R_0$, carrying capacity).
    \item \textbf{Structural target:} $y_{\text{struct}} = T(\mathcal{\theta})$, where $T$ assigns a label to the underlying model (e.g.\ “SIR’’ vs.\ “SEIR’’).
\end{enumerate}

\begin{definition}[Unobservable target]
A target, parametric or structural, is unobservable if, in real-world data, $x$ is never paired with its ground-truth value.
\end{definition}

\subsection{Simulation Enables Learning}

\begin{remark}[SGNNs learn the unobservable]\label{prop:unobservable}
Let $T(\theta)$ and $S(\mathcal{M})$ be unobservable targets.
\begin{enumerate}
    \item A learner trained on real-world pairs $\{(x_i,y'_i)\}$, where each $y'_i$ is \emph{observed}, receives no signal about $T(\theta)$ and cannot estimate it non-trivially.
    \item An SGNN trained on synthetic pairs $\bigl(x_i,T(\theta_i)\bigr)$ converges (Theorem~\ref{thm:finite-sample}) to the Bayes-optimal estimators
    \[
        f^*(x)=\mathbb{E}_{D_{\text{syn}}}[T(\theta)\!\mid\!x]
    \]
\end{enumerate}
\end{remark}

\emph{Proof sketch.}  Part (1) follows because the training loss never references the unobservable.  
For part (2), SGNNs create their own labelled data: draw $\theta\!\sim\!P(\theta)$, record the associated label, simulate $x$, then train by MSE or cross-entropy for structural targets.  By consistency (Prop.~\ref{prop:consistency}) the network converges to the conditional expectation, which is Bayes-optimal. \qed

\paragraph{Identifiability.}
Proposition~\ref{prop:unobservable} presumes that the target is identifiable for all candidate models:
\[
p(x\mid\theta)=p(x\mid\theta')\;\Longrightarrow\;T(\theta)=T(\theta').
\]
If this condition does not hold, then the target is not identifiable from the data distribution, and no method---regardless of model capacity or access to training data---can recover it.

\subsection{Illustrative Examples}

\begin{itemize}
    \item \textbf{Epidemiology — Parametric.}  SGNNs map noisy case trajectories to hidden parameters such as $R_0$ or incubation period, which are unmeasurable in real time.
    \item \textbf{Epidemiology — Structural.}  Given incidence curves, an SGNN can classify which compartmental model (SIR, SEIR, SIRS) best explains the data—information not present in the observations alone.
\end{itemize}

These examples highlight a key advantage of simulation-grounded learning: it turns scientifically meaningful but unobservable quantities into learnable targets, extending predictive modelling beyond the limits of traditional supervised learning.

\subsection{Experiment: Learning Unobserved Structural Targets}
\label{subsec:structural_experiment}

We empirically validate the theoretical prediction that SGNNs can learn unobservable structural targets by testing their ability to distinguish between competing epidemiological model structures. This experiment demonstrates both the practical utility of learning structural unobservables and provides concrete evidence that SGNNs outperform classical statistical approaches on this fundamental scientific task.

\paragraph{Experimental Setup.}
We consider the problem of distinguishing between SIR and SEIR epidemiological models from noisy trajectory observations. We implement discrete-time versions of both models with unit population. For SIR: $S_{t+1} = S_t - \beta S_t I_t$, $I_{t+1} = I_t + \beta S_t I_t - \gamma I_t$, $R_{t+1} = R_t + \gamma I_t$. For SEIR, we add an exposed compartment with transition rate $\sigma$. We generate 60,000 synthetic epidemics (30,000 each) by sampling $\beta \sim \mathcal{U}(0.1, 0.5)$, $\gamma \sim \mathcal{U}(0.05, 0.2)$, and for SEIR, $\sigma \sim \mathcal{U}(0.1, 0.3)$. Each trajectory runs for 100 time steps with additive Gaussian noise ($\sigma = 0.01$) applied to the infected compartment.

\paragraph{SGNN Architecture and Training.}
We train a 1D convolutional neural network with residual blocks, GroupNorm, and GELU activations to classify 100-dimensional infected time series as SIR (0) or SEIR (1). The architecture uses progressive channel expansion (64$\rightarrow$128$\rightarrow$256) with adaptive pooling and a fully connected classifier. Training uses cross-entropy loss with Adam optimizer (lr=$10^{-5}$) for 20 epochs on 48,000 samples.

\paragraph{AIC Baseline.}
For comparison, we implement AIC-based model selection by fitting both SIR and SEIR models to each test trajectory using nonlinear least squares, then selecting the model with lower AIC score: $\text{AIC} = n \log(\text{RSS}/n) + 2k$ where $k$ is the number of parameters \cite{aic}.

\begin{figure}[ht]
\centering
\includegraphics[width=0.6\textwidth]{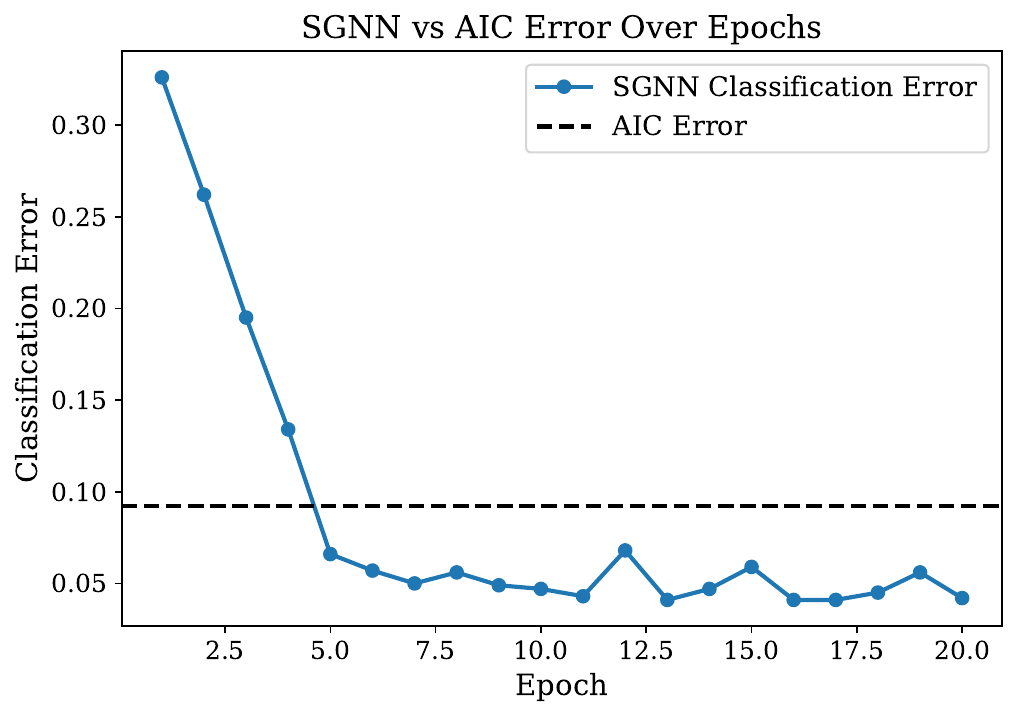}
\caption{
   \textbf{SGNNs outperform AIC for structural model selection.} Classification error over training epochs for distinguishing SIR vs SEIR models from noisy trajectory data. SGNNs rapidly converge to low error rates while AIC maintains significantly higher error, demonstrating the advantage of simulation-grounded learning for unobservable structural targets.
}
\label{fig:structural_comparison}
\end{figure}

\paragraph{Results.}
Figure~\ref{fig:structural_comparison} shows the results on 12,000 held-out samples. SGNNs achieve substantially lower classification error than AIC throughout training, converging to approximately 4\% error while AIC maintains 9.2\% error. This demonstrates that SGNNs can reliably learn structural differences between mechanistic models that classical statistical methods struggle to distinguish.

\paragraph{Interpretation.}
This experiment validates Proposition~\ref{prop:unobservable} for structural targets. SGNNs learn robust mappings from noisy trajectories to mechanistic structure through simulation-based training, while AIC struggles with parameter estimation uncertainty. This demonstrates the practical utility of simulation-grounded learning for scientific model selection.

\section{Discussion}

This work presents a formal foundation for simulation-grounded prediction, framing Simulation-Grounded Neural Networks (SGNNs) as consistent approximators of the Bayes-optimal predictor under a simulation-induced data distribution. The theoretical results unify several previously ad hoc insights about SGNN behavior, yielding interpretable guarantees and failure modes.

A central result of this theory is that SGNNs trained on synthetic data perform amortized Bayesian estimation: they learn to map observed data to the posterior expectation of a scientific quantity of interest under the simulator’s generative model. This quantity may be the latent mechanism itself (e.g., simulator parameters) or a downstream function of it (e.g., reproduction number, source node, or model structure). In both cases, the SGNN approximates the Bayes-optimal predictor $f^*(x) = \mathbb{E}[y \mid x]$. This perspective provides a principled foundation for understanding generalization, consistency, and interpretability in simulation-grounded learning (Section~\ref{sec:sgns-bayes}).

Our analysis of generalization under model misspecification (Section~\ref{sec:mismatch}) adapts classical distribution-shift bounds to the SGNN setting, decomposing test risk into two terms: synthetic excess risk (arising from limited simulation samples or model capacity) and a mismatch penalty (arising from divergence between the simulator and reality). This decomposition formalizes the robustness properties observed in practice and highlights how simulator design influences downstream generalization.

We also formalized two distinctive aspects of SGNNs. First, we showed that back-to-simulation attribution can recover posterior information about latent mechanisms, and under alignment training converges in all moments to the true posterior (Section~\ref{sec:b2s}). Second, we demonstrated that SGNNs can estimate unobservable scientific targets $T(\theta)$ whenever they are identifiable from simulated data (Section~\ref{sec:unobservable}). These results illustrate how simulation-grounded learning extends beyond standard supervised paradigms by turning scientifically meaningful but unobserved quantities into estimable ones.

\subsection{Implications for Scientific Machine Learning}

The theoretical framework developed here has direct implications for scientific model design and deployment.

\paragraph{The Simulator as an Inductive Bias.} Rather than treating a simulator as a fixed model of the world, our formulation interprets it as a structured prior over hypotheses, encoded through distributions over parameters, mechanisms, and observation processes. Training an SGNN corresponds to amortizing inference over this space, enabling prediction even when empirical data is limited or incomplete.

\paragraph{Tradeoffs in Simulation Fidelity.} The generalization bound in Theorem~\ref{thm:mismatch} formalizes a trade-off between simulator complexity and learnability. Oversimplified simulators yield a large mismatch penalty, while overparameterized simulators may yield high synthetic risk due to variance and complexity. Effective simulator design involves identifying a regime where the synthetic task is both expressive and statistically learnable.

\paragraph{Data-Limited Regimes.} In domains where empirical data is scarce or unlabelable, simulation-grounded learning offers a viable alternative. The ability to learn from unobservable quantities that are defined mechanistically but not measured extends the reach of predictive modeling in scientific contexts where traditional labels are unavailable.

\subsection{Limitations and Future Directions}

While this work establishes foundational results, several directions remain for refinement and extension.

\paragraph{Mismatch Estimation and Control.} The mismatch penalty depends on divergences such as $\Delta_{TV}(D_{\text{real}} \,\|\, D_{\text{syn}})$, which are generally unobservable. Estimating or bounding this term from unlabeled real-world data, possibly using domain adaptation or two-sample testing methods, is an important direction for enabling practical performance guarantees.

\paragraph{Active Simulation Design.} Our framework assumes a fixed simulator prior $P(\theta)$. In practice, simulation may be expensive. Incorporating active learning or adaptive simulation strategies could allow more efficient coverage of the hypothesis space, reducing sample complexity and improving generalization.

\paragraph{Beyond Prediction: Causality and Control.} The SGNN paradigm naturally extends to interventional and dynamic settings. For instance, SGNNs could be trained on interventional simulators to estimate treatment effects or learn policies under uncertainty. Establishing consistency and generalization guarantees in these contexts is a promising direction for both theory and application.

\section{Conclusion}

We develop a formal theory of simulation-grounded prediction and analyze Simulation-Grounded Neural Networks (SGNNs) as approximators of the Bayes-optimal predictor under a structured simulation prior. Our results adapt classical learning theory to the SGNN setting, yielding a generalization bound that isolates the effect of simulator–reality mismatch, conditions under which unobservable scientific targets become learnable, and convergence guarantees for attribution methods trained with alignment objectives. Together, these findings provide rigorous foundations for SGNNs and clarify both their capabilities and their limitations in data-limited scientific domains where traditional empirical learners struggle. More broadly, this work bridges mechanistic modeling and modern machine learning, pointing toward a principled path for interpretable, robust, and scientifically grounded prediction.

\bibliographystyle{unsrt}

\begin{thebibliography}{10}

\bibitem{sgnns}
Carson Dudley, Reiden Magdaleno, Christopher Harding, and Marisa Eisenberg.
\newblock Simulation as supervision: Mechanistic pretraining for scientific discovery.
\newblock {\em arXiv preprint arXiv:2507.08977}, 2025.

\bibitem{mantis}
Carson Dudley, Reiden Magdaleno, Christopher Harding, Ananya Sharma, Emily Martin, and Marisa Eisenberg.
\newblock Mantis: A simulation-grounded foundation model for disease forecasting.
\newblock {\em arXiv preprint arXiv:2508.12260}, 2025.

\bibitem{mohri2018foundations}
Mehryar Mohri, Afshin Rostamizadeh, and Ameet Talwalkar.
\newblock {\em Foundations of Machine Learning}.
\newblock MIT Press, 2nd edition, 2018.

\bibitem{identifiability1}
Marisa~C Eisenberg, Suzanne~L Robertson, and Joseph~H Tien.
\newblock Identifiability and estimation of multiple transmission pathways in cholera and waterborne disease.
\newblock {\em Journal of Theoretical Biology}, 1 2013.

\bibitem{identifiability2}
Marisa~C Eisenberg and Michael~AL Hayashi.
\newblock Determining identifiable parameter combinations using subset profiling.
\newblock {\em Mathematical Biosciences}, 256:116--126, 10 2014.

\bibitem{identifiability3}
Yu-Han Kao and Marisa~C Eisenberg.
\newblock Practical unidentifiability of a simple vector-borne disease model: Implications for parameter estimation and intervention assessment.
\newblock {\em Epidemics}, 25:89--100, 12 2018.

\bibitem{cranmer2020frontier}
Kyle Cranmer, Johann Brehmer, and Gilles Louppe.
\newblock The frontier of simulation-based inference.
\newblock {\em Proceedings of the National Academy of Sciences}, 2020.

\bibitem{raissi2019pinns}
Maziar Raissi, Paris Perdikaris, and George~Em Karniadakis.
\newblock Physics‑informed neural networks: A deep learning framework for solving forward and inverse problems involving nonlinear partial differential equations.
\newblock {\em Journal of Computational Physics}, 378:686--707, 2019.

\bibitem{neuraloperator}
Nikola Kovachki, Zongyi Li, Burigede Liu, Kamyar Azizzadenesheli, Kaushik Bhattacharya, Andrew Stuart, and Anima Anandkumar.
\newblock Neural operator: Learning maps between function spaces with applications to pdes.
\newblock {\em Journal of Machine Learning Research}, 2023.

\bibitem{uodes}
Christopher Rackauckas, Yingbo Ma, Julius Martensen, Collin Warner, Kirill Zubov, Rohit Supekar, Dominic Skinner, Ali Ramadhan, and Alan Edelman.
\newblock Universal differential equations for scientific machine learning, 2021.

\bibitem{defsi}
Lijing Wang, Jiangzhuo Chen, and Madhav Marathe.
\newblock Defsi: Deep learning based epidemic forecasting with synthetic information.
\newblock In {\em Proceedings of the AAAI Conference on Artificial Intelligence}, volume~33, 2019.

\bibitem{sim2real}
Abhishek Kadian, Joanne Truong, Aaron Gokaslan, Alexander Clegg, Erik Wijmans, Stefan Lee, Manolis Savva, Sonia Chernova, and Dhruv Batra.
\newblock Sim2real predictivity: Does evaluation in simulation predict real-world performance?
\newblock {\em IEEE Robotics and Automation Letters}, 5(4):6670–6677, October 2020.

\bibitem{pfn}
Samuel Müller, Noah Hollmann, Sebastian Pineda~Arango, Josif Grabocka, and Frank Hutter.
\newblock Transformers can do bayesian inference.
\newblock {\em arXiv preprint arXiv:2112.10510}, 2021.

\bibitem{tabpfn}
Noah Hollmann, Samuel Müller, Katharina Eggensperger, and Frank Hutter.
\newblock Tabpfn: A transformer that solves small tabular classification problems in a second, 2023.

\bibitem{nagler2023foundations}
Thomas Nagler.
\newblock Statistical foundations of prior‑data fitted networks.
\newblock In {\em International Conference on Machine Learning}, 2023.

\bibitem{bartlett2002rademacher}
Peter~L. Bartlett and Shahar Mendelson.
\newblock Rademacher and gaussian complexities: Risk bounds and structural results.
\newblock {\em Journal of Machine Learning Research}, 3:463--482, 2002.

\bibitem{aic}
H.~Akaike.
\newblock A new look at the statistical model identification.
\newblock {\em IEEE Transactions on Automatic Control}, 19(6):716--723, 1974.

\bibitem{sindy}
Steven~L. Brunton, Joshua~L. Proctor, and J.~Nathan Kutz.
\newblock Discovering governing equations from data by sparse identification of nonlinear dynamical systems.
\newblock {\em Proceedings of the National Academy of Sciences}, 113(15):3932--3937, 2016.

\bibitem{lassoode}
Jiale Tan and Marisa~C. Eisenberg.
\newblock Lasso-ode: A framework for mechanistic model identifiability and selection in disease transmission modeling, 2025.

\bibitem{shap}
Scott Lundberg and Su-In Lee.
\newblock A unified approach to interpreting model predictions, 2017.

\bibitem{lime}
Marco~Tulio Ribeiro, Sameer Singh, and Carlos Guestrin.
\newblock "why should i trust you?": Explaining the predictions of any classifier, 2016.

\bibitem{koh2020understanding}
Pang~Wei Koh and Percy Liang.
\newblock Understanding black-box predictions via influence functions, 2020.

\end{thebibliography}

\appendix
\section{Proofs of Main Results}
\label{app:proofs}

\subsection{Proof of Theorem~\ref{thm:finite-sample}}

\begin{proof}
\textbf{Step 1: Bias-variance decomposition.}
The total excess risk naturally decomposes as:
\begin{align}
R_{\text{syn}}(f_{\phi_N}) - R_{\text{syn}}(f^*) &= \left( R_{\text{syn}}(f_{\phi_N}) - \inf_{f \in \mathcal{F}} R_{\text{syn}}(f) \right) + \left( \inf_{f \in \mathcal{F}} R_{\text{syn}}(f) - R_{\text{syn}}(f^*) \right) \\
&= \mathcal{E}_{\text{est}} + \mathcal{E}_{\text{approx}}
\end{align}

The approximation error $\mathcal{E}_{\text{approx}}$ measures the fundamental limitation of our function class $\mathcal{F}$ and depends only on the expressiveness of the model family. Our goal is to bound the estimation error $\mathcal{E}_{\text{est}}$.

\textbf{Step 2: Define the population risk minimizer.}
Let $f_{\mathcal{F}}^* \in \arg\min_{f\in\mathcal{F}} R_{\text{syn}}(f)$ be the function in $\mathcal{F}$ that achieves the population risk infimum. Note that $R_{\text{syn}}(f_{\mathcal{F}}^*) = \inf_{f\in\mathcal{F}} R_{\text{syn}}(f)$.

\textbf{Step 3: Rademacher concentration inequality.}
We apply the following generalization of Mohri et al.\ (2018), Theorem 3.3:

\begin{lemma}[Concentration for bounded function classes]
For a function class $\mathcal{G}$ with functions $g: \mathcal{Z} \to [0,B]$, with probability at least $1-\delta$:
\begin{equation}
\sup_{g \in \mathcal{G}} \left( R(g) - \widehat{R}_N(g) \right) \leq 2\,\widehat{\mathfrak{R}}_N(\mathcal{G}) + 3B\sqrt{\frac{\log(2/\delta)}{2N}} \label{eq:concentration}
\end{equation}
where $R(g) = \mathbb{E}[g]$ and $\widehat{R}_N(g) = \frac{1}{N}\sum_{i=1}^N g(z_i)$.
\end{lemma}

\textbf{Step 4: Apply to loss class.}
Define the loss class $\mathcal{G} = \{z \mapsto \ell(f(z), y) : f \in \mathcal{F}\}$. Since $\ell$ maps to $[0,B]$, each function in $\mathcal{G}$ is bounded by $B$.

Applying the concentration inequality \eqref{eq:concentration} to $g_f(z) = \ell(f(z), y)$ gives us, with probability $1-\delta$:
\begin{equation}
R_{\text{syn}}(f) - \widehat{R}_N(f) \leq 2\,\widehat{\mathfrak{R}}_N(\mathcal{G}) + 3B\sqrt{\frac{\log(2/\delta)}{2N}} \quad \forall f \in \mathcal{F} \label{eq:uniform_bound}
\end{equation}

\textbf{Step 5: Apply to specific functions.}
From \eqref{eq:uniform_bound}, we have:
\begin{align}
R_{\text{syn}}(f_{\phi_N}) &\leq \widehat{R}_N(f_{\phi_N}) + 2\,\widehat{\mathfrak{R}}_N(\mathcal{G}) + 3B\sqrt{\frac{\log(2/\delta)}{2N}} \label{eq:bound_phi} \\
R_{\text{syn}}(f_{\mathcal{F}}^*) &\leq \widehat{R}_N(f_{\mathcal{F}}^*) + 2\,\widehat{\mathfrak{R}}_N(\mathcal{G}) + 3B\sqrt{\frac{\log(2/\delta)}{2N}} \label{eq:bound_fstar}
\end{align}

\textbf{Step 6: Use ERM property.}
By definition of empirical risk minimization: $\widehat{R}_N(f_{\phi_N}) \leq \widehat{R}_N(f_{\mathcal{F}}^*)$.

\textbf{Step 7: Combine bounds.}
Subtracting \eqref{eq:bound_fstar} from \eqref{eq:bound_phi}:
\begin{align}
R_{\text{syn}}(f_{\phi_N}) - R_{\text{syn}}(f_{\mathcal{F}}^*) &\leq \left(\widehat{R}_N(f_{\phi_N}) - \widehat{R}_N(f_{\mathcal{F}}^*)\right) + 2 \cdot \left[2\,\widehat{\mathfrak{R}}_N(\mathcal{G}) + 3B\sqrt{\frac{\log(2/\delta)}{2N}}\right] \\
&\leq 0 + 4\,\widehat{\mathfrak{R}}_N(\mathcal{G}) + 6B\sqrt{\frac{\log(2/\delta)}{2N}}
\end{align}

\textbf{Step 8: Apply Talagrand's contraction lemma.}
Since $\ell$ is $L$-Lipschitz in its first argument, Talagrand's Contraction Lemma (Mohri et al., Lemma 5.7) gives:
\[
\widehat{\mathfrak{R}}_N(\mathcal{G}) \leq L\,\widehat{\mathfrak{R}}_N(\mathcal{F})
\]

Therefore:
\[
\mathcal{E}_{\text{est}} = R_{\text{syn}}(f_{\phi_N}) - R_{\text{syn}}(f_{\mathcal{F}}^*) \leq 4L\,\widehat{\mathfrak{R}}_N(\mathcal{F}) + 6B\sqrt{\frac{\log(2/\delta)}{2N}}
\]

\textbf{Step 9: Final result.}
Combining with the approximation error:
\[
R_{\text{syn}}(f_{\phi_N}) - R_{\text{syn}}(f^*) \leq \left(\inf_{f\in\mathcal{F}}R_{\text{syn}}(f) - R_{\text{syn}}(f^*)\right) + 4L\,\widehat{\mathfrak{R}}_N(\mathcal{F}) + 6B\sqrt{\frac{\log(2/\delta)}{2N}}
\]
\end{proof}

\subsection{Proof of Theorem~\ref{thm:mismatch}}

\begin{proof}
The core of the proof relies on relating the risks under $D_{real}$ to those under $D_{syn}$. For any predictor $f$, the difference in its expected loss under the two distributions is bounded by the Total Variation distance:
\begin{equation}
    |R_{real}(f) - R_{syn}(f)| = |\mathbb{E}_{D_{real}}[\ell(f(x),y)] - \mathbb{E}_{D_{syn}}[\ell(f(x),y)]| \le L_{max} \Delta_{TV}. \label{eq:tv_bound}
\end{equation}
This gives us two useful inequalities:
\begin{align}
    R_{real}(f) &\le R_{syn}(f) + L_{max}\Delta_{TV} \label{eq:ineq1} \\
    -R_{real}(f) &\le -R_{syn}(f) + L_{max}\Delta_{TV}. \label{eq:ineq2}
\end{align}
We now bound the excess risk on the real distribution.
\begin{align*}
    R_{real}(f_{\phi_N}) - R_{real}(f^{\star}_{real}) & \le \left( R_{syn}(f_{\phi_N}) + L_{max}\Delta_{TV} \right) - R_{real}(f^{\star}_{real}) \quad &&\text{by \eqref{eq:ineq1} for } f_{\phi_N} \\
    & \le R_{syn}(f_{\phi_N}) + L_{max}\Delta_{TV} + \left( -R_{syn}(f^{\star}_{real}) + L_{max}\Delta_{TV} \right) \quad &&\text{by \eqref{eq:ineq2} for } f^{\star}_{real} \\
    & = R_{syn}(f_{\phi_N}) - R_{syn}(f^{\star}_{real}) + 2 L_{max} \Delta_{TV}.
\end{align*}
By definition, $f^{\star}_{syn}$ is the minimizer of the risk under $D_{syn}$, so $R_{syn}(f^{\star}_{syn}) \le R_{syn}(f^{\star}_{real})$. This implies $-R_{syn}(f^{\star}_{real}) \le -R_{syn}(f^{\star}_{syn})$. Substituting this into the last line gives the final bound:
\[
R_{real}(f_{\phi_N}) - R_{real}(f^{\star}_{real}) \le R_{syn}(f_{\phi_N}) - R_{syn}(f^{\star}_{syn}) + 2 L_{max} \Delta_{TV}. \qedhere
\]
\end{proof}

\subsection{Proof of Theorem~\ref{thm:bsa_consistency}}

\begin{proof}
\textbf{Step 1: Prediction consistency.} 
From Theorem~\ref{thm:finite-sample}, $f_\phi(x) \to \mathbb{E}[T(\theta) \mid x]$.

\textbf{Step 2: KL convergence.} 
Since $\mathcal{L}_{\text{KL}} \to 0$, we have $\text{KL}(p(\theta \mid x) \| \sum_i w_i(x) \delta_{\theta_i}) \to 0$ for almost every $x$.

\textbf{Step 3: Moment convergence via KL convergence.}
For any bounded measurable function $g: \Theta \to \mathbb{R}$, KL convergence implies:
\[
\left|\mathbb{E}_{p(\theta|x)}[g(\theta)] - \sum_{i=1}^M w_i(x) g(\theta_i)\right| \to 0
\]

This is because KL divergence controls the total variation distance, and total variation convergence implies convergence of expectations for all bounded functions.

\textbf{Step 4: Apply to moment functions.}
For any positive integer $k$, define $g_k(\theta) = T(\theta)^k$. Since $T(\theta)$ is bounded, $g_k(\theta)$ is also bounded. Applying Step 3:
\[
\sum_{i=1}^M w_i(x) T(\theta_i)^k = \sum_{i=1}^M w_i(x) g_k(\theta_i) \to \mathbb{E}_{p(\theta|x)}[g_k(\theta)] = \mathbb{E}[T(\theta)^k \mid x]
\]

\end{proof}

\end{document}